# Automatic diagnosis of retinal diseases from color retinal images

D.Jayanthi
PG Scholar
Dept of CSE
Sri Venkateswara college
of Engineering

N.Devi
Senior Lecturer
Dept of IT
Sri Venkateswara college
of Engineering

S.SwarnaParvathi
Senior Lecturer
Dept of IT
Sri Venkateswara college
of Engineering

*Abstract*-**Teleophthalmology holds a great potential to improve the quality, access, and affordability in health care. For patients, it can reduce the need for travel and provide the access to a super-specialist. Ophthalmology lends itself easily to telemedicine as it is a largely image based diagnosis. The main goal of the proposed system is to diagnose the type of disease in the retina and to automatically detect and segment retinal diseases without human supervision or interaction. The proposed system will diagnose the disease present in the retina using a neural network based classifier.The extent of the disease spread in the retina can be identified by extracting the textural features of the retina. This system will diagnose the following type of diseases: Diabetic Retinopathy and Drusen.**

## I. INTRODUCTION

The World Health Organization estimates that 135 million people have diabetes mellitus worldwide and that the number of people with diabetes will increase to 300 million by the year 2025.Early detection and treatment of these diseases are crucial to avoid preventable vision loss.

Diabetes mellitus (DM) is a chronic, systemic, life-threatening disease characterized by disordered metabolism and abnormally high blood sugar (hyperglycemia) resulting from low levels of the hormone insulin with or without abnormal resistance to insulin's effects. Through computer simulations it is possible to demonstrate that prevention and treatment are relatively inexpensive compared to the healthcare and rehabilitation costs incurred by visual loss or blindness.

Assessment of the risk for development of age-related macular degeneration (ARMD) requires reliable detection and quantitative mapping of retinal abnormalities that are considered as precursors of the disease. Typical signs for the latter are the so-called drusen that appear as abnormal white-yellow deposits on the retina. Color retinal images are used presently to visually identify the presence of drusens. Segmentation of these features using conventional image analysis methods is quite complicated mainly due to the non-uniform illumination and the variability of the pigmentation of the background tissue. Automated detection and analysis can provide vital information about the quantity and quality of the drusens.

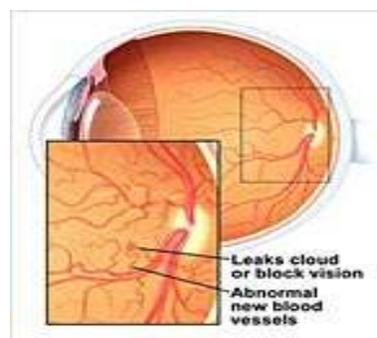

Fig.1 Diabetic Retinopathy

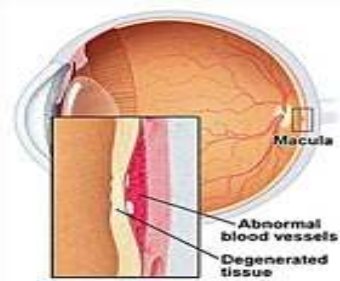

Fig.2 Age-related Macular Degeneration

## II. PROPOSED SYSTEM

In the proposed system, a combined method for classifying the type of retinal disease and automatic diagnosis can be done using a neural network classifying technique. The main goal of the proposed system is to diagnose the age-related macular degeneration (drusen) and diabetic retinopathy diseases in the retina and to automatically detect and segment the above diseases without human supervision or interaction. Texture analysis is used to extract the features of the retina. After feature extraction process, a neural network based classifier is used to classify the type of retinal disease and can automatically diagnose the type of disease. The location of drusen and diabetes's can be identified and by extracting the textural features of the retina, the extent of disease spread can be determined.

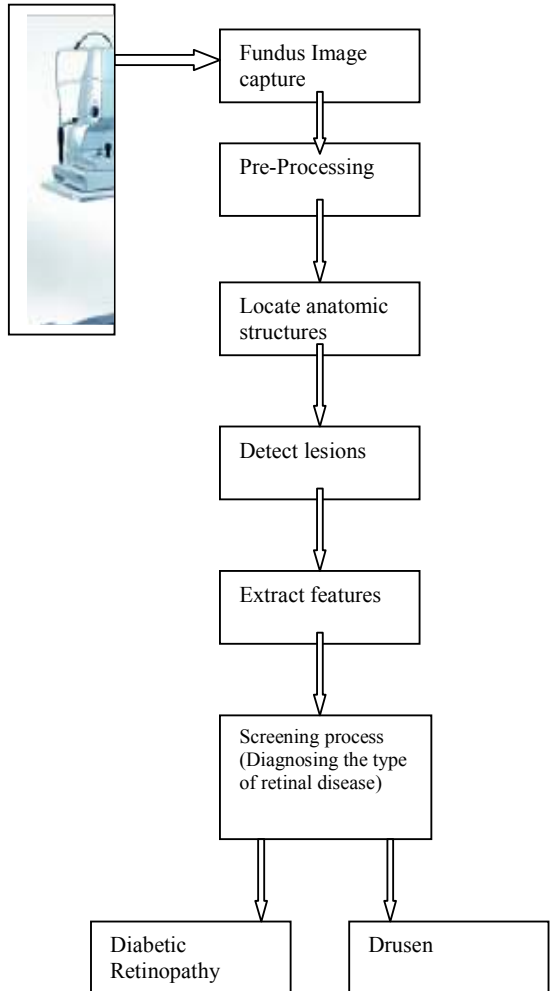

Fig. 3 Block diagram of the proposed system

### III. MODULES

In the proposed system, there are four modules, they are as follows:
1. Pre-Processing of a color retinal image.
2. Locating anatomic structures and detecting lesions.
3. Feature Extraction.
4. Classification of retinal disease using Artificial Neural Networks.

### A. Pre-Processing

Pre-Processing of retinal image is the first step in the automatic diagnosis of retinal diseases. The problem with retinal image is that the quality of the acquired images is usually not good. So, it is necessary to improve the quality of retinal image. The purpose of pre-processing is to remove the noisy area from retinal image. This is required for the reliable extraction of features and abnormalities as feature extraction and abnormality detection algorithms give poor results in the presence of noisy background..

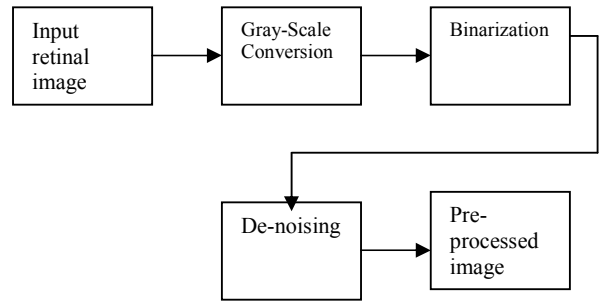

Fig. 4 Block diagram for preprocessing

*Steps for pre-processing*

1. Divide the input retinal image into non-overlapping blocks.
2. Extract RGB components from the original color retinal image.
3. After gray-level conversion, use histogram equalization to enhance the contrast and to improve the quality of retinal image.
4. Use a large median filter to remove the noise from the image.

### B. Locating anatomic structures and detecting lesions

The purpose of locating anatomic structures is to detect the optic nerve based on segmentation of the vascular arcades. Detection of the anatomic structures is fundamental to the subsequent characterization of the normal or disease state that may exists in the retina. The algorithm is based on mathematical morphology and curvature evaluation for the detection of vessel like patterns in a noisy environment. Vessel detection is based on the computation of parameters related to blood flow. In order to define the vessel like pattern, segmentation will be performed with respect to a precise model. In order to differentiate vessels from analogous patterns, a cross curvature evaluation is performed. Vessel like patterns are bright features defined by morphological properties like linearity, connectivity, width and by a specific Gaussian like profile whose curvature varies smoothly along the vessel.

The detection algorithm is based on four steps.
1. Noise Reduction.
2. Linear pattern with Gaussian-like profile improvement.
3. Cross curvature evaluation.
4. Linear filtering.

Our goal is to produce a binary image of the vasculature, b( i , j), for an image of size I x J. We want to achieve a robust segmentation of for a wide variety of images representing various states of the retinal disease.

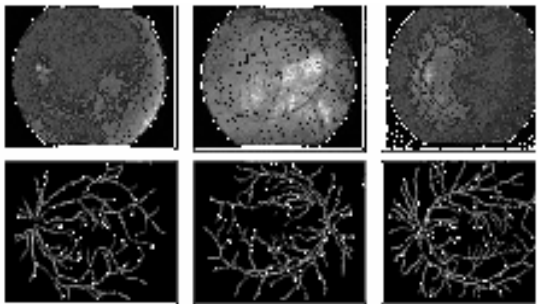

Fig.5 Vascular Segmentation with age-related macular degeneration (drusen)

*Detection of Blood Vessels Boundaries*
*Algorithm 1 – Boundary detection using image statistics*

Image statistics such as mean and standard deviation can be used in boundary detection. The algorithm called DBDED, which stands for decision-based directional edge detector uses image statistics. It is used to detect the boundaries of blood vessel tree in the retinal images. Each point that passes a local threshold is investigated as a an edge candidate. Then, the point (x , y) is to be a one-dimensional edge candidate from the east direction if it satisfies the following equation:

$$I(x,y) \geq Av[2I(x+1,y), I(x+2,y), I(x+3,y)] + Sd[2I(x+1,y), I(x+2,y), I(x+3,y)] + \eta$$

Where I(x, y) is the intensity of an image at coordinates x and y, $\eta$ is a constant threshold and $Av(.)$ and $Sd(.)$ denote average and standard deviation operations respectively

The decision algorithm is applied and the points are considered as 2-D edges if they satisfy the following conditions
1. They are 1-D Edge candidates in at least two and at most seven directions.
2. If (x, y) is an edge candidate then, at least one of the immediate 8-neighboring points:
$$\{(x+\beta, y+\gamma) \mid \beta \in (0,1,-1), \gamma \in (0,1,-1), \beta + \beta\gamma + \gamma \neq 0\}$$
is also an edge candidate.

*Algorithm 2 – Extraction of blood vessel boundaries using deformable models*

The recent one of the methods of contour detection is deformable models are snake. A snake is an active contour model that is manually initiated near to the contour of interest. This contour model deforms according to some criteria and image features to finally stay to the actual contour(s) in the image. An energy function is formulated to obtain an estimate of the quality of the mode in terms of its internal shape, and external forces e.g. underlying image forces and user constraint forces. The energy function integrates the weighted linear combination of the internal and external forces of the contour.

*Extraction of the core area of the blood vessel tree by tracing vessel centers*

*Algorithm : Extraction of blood vessel tree using the morphological reconstruction*

Morphological reconstruction is to reconstruct an object in an image, called the marker image, containing at least one point belonging to that object from an image, called the mask image, containing that object and other objects and noise. An efficient implementation of morphological reconstruction can be described as follows:
1. Label the connected components of the mask image, (i.e.) each of these components is assigned a unique number.
2. Determine the labels of the connected components, which contain at least a pixel of the marker image.
3. Remove all the connected components that are not of the previous ones.

C. FEATURE EXTRACTION

The aim of texture analysis is texture recognition and texture based shape analysis. The texture can be studied in two levels namely statistical and structural. On the statistical level, the texture of an image is defined by a set of statistics extracted from the entire texture region. On the structural level, a texture is defined by sub patterns called primitives. The various statistical methods are based on capturing the variability in grey scale images. The textural character of an image depends on the spatial size of texture primitives. Large primitives give rise to coarse texture (e.g. rock surface) and small primitives gives fine texture (e.g. silk surface).
Feature extraction can be done in two steps.
1. Features detecting optic nerve.
2. Features detecting diseases.

*Features detecting optic nerve*

Applying the pre-processing techniques like noise removal and histogram equalization, we obtain a better contrast image with the different objects having different textures. Each image is divided into large number of equal parts and local mean, standard deviation and variance are calculated. The characteristics of vessel structure are,

1. *Retina luminancel(i,j)* -This feature measures the brightness that can be helpful for locating the retinal lesions

$$\ell(i,j) = \frac{1}{M \cdot N} \sum_{m=0}^{M-1} \sum_{n=0}^{N-1} I(i-m, j-m)$$

which supports M X N region for every point ( i , j ) in the image.

2. *Vessel density, $\rho(i,j)$* -Vessel density is defined as the number of vessels existing in a unit area of the retina. Since the vasculature that feeds the retina enters the eye, the vessels tend to be most dense in this region.

$$\rho(i,j) = \frac{1}{M \cdot N} \sum_{m=0}^{M-1} \sum_{n=0}^{N-1} b_r(i-m, j-n),$$

which supports M X N region for every point ( i , j ) in the image.

3. *Average vessel thickness, $t(i,j)$* -Vessels are also observed to be thickest near the optic nerve since most branching of both the arterial and venous structures does not take place until the tree is more distal from the optic nerve.

$$t(i,j) = \frac{\sum_{m=0}^{M-1} \sum_{n=0}^{N-1} b_r(i-m, j-n)}{\sum_{m=0}^{M-1} \sum_{n=0}^{N-1} b_t(i-m, j-n)},$$

which supports M X N region for every point ( i , j ) in the image.

4. *Average Vessel Orientation, $\theta(i,j)$* -The vessels entering the eye are roughly perpendicular to the horizontal raphe of the retina. i.e. the demarcation line running through the optic nerve and fovea. The result is an observation of vascular orientation being ±90 ° relative to the horizontal raphe when entering the eye and becoming more parallel (i.e., 0°) as the distance from the optic nerve increases.

$$\theta(i,j) = \frac{1}{M \cdot N} \sum_{m=0}^{M-1} \sum_{n=0}^{N-1} b_r(i-m, j-n) \cdot \cos\theta(i-m, j-n).$$

which supports M X N region for every point ( i , j ) in the image.

*Features detecting retinal diseases*

Statistical texture features describe texture in a form, which is suitable for pattern recognition. As a result each texture is described by a feature vector of properties, which represents a point in a multi dimensional feature space.

a) *Mean*-The nth moment of about the mean is

$$\mu_n(z) = \sum_{i=0}^{L-1} (z_i - m)^n p(z_i) \quad (1)$$

where m is the **mean value** of z (the average gray level) :

$$m = \sum_{i=0}^{L-1} z_i \, p(z_i)0$$

Note from Eq. (1) that $\mu_0 = 1$ and $\mu_1 = 0$.

b) *Variance*-The second moment [the **variance** $\sigma^2(z) = \mu_2(z)$] is of particular importance in texture description. It is a measure of gray-level contrast that can be used to establish descriptors of relative smoothness.

c) *Skewness*- The third moment,

$$\mu_3(z) = \sum_{i=0}^{L-1} (z_i - m)^3 p(z_i)$$

is a measure of the **skewness** of the histogram.

d) *Entropy*

$H(z) = -\int p(z) \ln p(z) \, dz$ is Shannon's entropy of the image window **z**, and *p* is the distribution of the grey levels in the considered window. We approximate the entropy as:

$$H(z) \approx -\frac{1}{N_z} \sum_{q_i \in z} \ln \frac{1}{N_p} \sum_{ij \in W_p} g_\psi(z_i - z_j)$$

e) *Correlation distance*-The distances among all the images can be computed using correlation distance.

$$d_{rs} = 1 - \frac{(x_r - \bar{x}_r)(x_s - \bar{x}_s)^T}{\left[(x_r - \bar{x}_r)(x_r - \bar{x}_r)^T\right]^{\frac{1}{2}} \left[(x_s - \bar{x}_s)(x_s - \bar{x}_s)^T\right]^{\frac{1}{2}}},$$

where $\bar{x}_r = \frac{1}{n} \sum_j x_{rj}$ and $\bar{x}_s = \frac{1}{n} \sum_j x_{sj}$.

g) *Zernike moments (ZM)*- The complex zernike moments of order n with repetition *l* are defined as

$$A_{nl} = \frac{n+1}{\pi} \int_0^{2\pi} \int_0^\infty [V_{nl}(r,\theta)]^* \cdot f(r\cos\theta, r\sin\theta) r \, dr \, d\theta$$

Where n = 0,1,2,...., and l takes on positive and negative integer values subject to the conditions $n - |l| = $ even, $|l| \le n$. The symbol * denotes complex conjugate. The zernike polynomials

$$v_{nl}(x,y) = V_{nl}(r\cos\theta, r\sin\theta) = R_{nl}(r) e^{il\theta}$$

are a complete set of complex-valued functions orthogonal to the unit disk.

The function f(x, y) can be expanded in terms of zernike polynomials over the unit disk as

$$f(x, y) = \sum_{n=0}^{\infty} \sum_{\substack{l=-\infty \\ n-|l|=\text{even} \\ |l| \le n}}^{\infty} A_{nl} V_{nl}(x, y)$$

*D. Classification of retinal disease using Artificial Neural Networks*

Artificial Neural Networks (ANN) has been used in a number of different ways in medicine and medically related fields. The principal advantage of ANN is to generalize, adapting to signal distortion and noise without the loss of robustness. Auto Associative Neural Network (AANN) is a network having the same number of neurons in input and output layers, and the less in the hidden layers. The network is trained using the input vector itself as the desired output. This training leads to organize a compression/encoding network between the input layer and the hidden layer, and a decoding layer between the hidden layer and output layer. Each of the autoassociative networks is trained independently for each class using the feature vector of the class. The squared error between an input and the output is generally minimized by the network of the class to which the input pattern belongs. This property enables us to classify an unknown input pattern. The unknown pattern is fed to all the networks, and is classified to the class with minimum squared error.

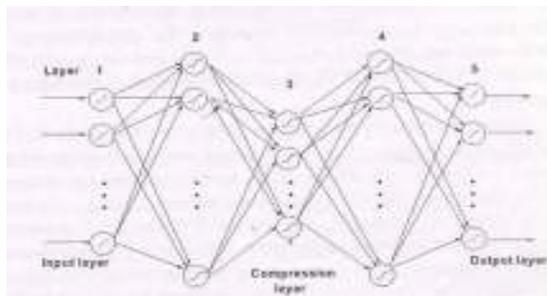

Fig 5 Auto Associative Neural Network

IV. CONCLUSION

Computer aided diagnosis of retinal diseases is one of the most important tasks when dealing with a huge population. This paper gives a survey of the classical and up-to-date methods for classifying and diagnosing the type of retinal disease and detecting its features after diagnosis at an earlier stage of the disease. Although a lot of work has been done, automatic diagnosis of retinal diseases at an earlier stage still remains an open problem.

The paper gives only the frame work for diagnosing human retinal diseases. This can be implemented using matlab. Each module can be tested individually with a test data of size 100. The results can be classified into four phases: true positive, true negative, false positive, false negative. The major goal of the paper is to provide a comprehensive reference source for the researchers involved in automatic diagnosis of retinal images. This framework can be extended to any number of retinal diseases in future.